\documentclass[lettersize,journal]{IEEEtran}
\usepackage{amsmath,amsfonts}
\usepackage{algorithmic}
\usepackage{algorithm}
\usepackage{array}
\usepackage[caption=false,font=normalsize,labelfont=sf,textfont=sf]{subfig}
\usepackage{textcomp}
\usepackage{stfloats}
\usepackage{url}
\usepackage{verbatim}
\usepackage{graphicx}
\usepackage{cite}

\usepackage[table]{xcolor} 
\usepackage{tabularx} 
\usepackage{booktabs} 
\usepackage{amsmath} 
\usepackage{multirow}
\usepackage{booktabs}
\usepackage{siunitx}
\usepackage{adjustbox}
\usepackage{changepage}

\usepackage{colortbl} 
\definecolor{lightgray}{gray}{0.93}
\definecolor{darkgray}{gray}{0.85}

\begin{document}

\title{Decoupled Complementary Spectral–Spatial Learning for Background Representation Enhancement in Hyperspectral Anomaly Detection}

\author{Wenping Jin\textsuperscript{1}, Li Zhu\textsuperscript{1,*}, Fei Guo\textsuperscript{1}
	\thanks{\textsuperscript{1}School of Software Engineering, Xi'an Jiaotong University, Xi'an 710049, China. E-mails: jinwenping@stu.xjtu.edu.cn, zhuli@xjtu.edu.cn, co.fly@stu.xjtu.edu.cn}
	\thanks{*Corresponding author}}

\markboth{}%
{Shell \MakeLowercase{\textit{et al.}}: A Sample Article Using IEEEtran.cls for IEEE Journals}

\IEEEpubid{}


\maketitle

\begin{abstract}

A recent class of hyperspectral anomaly detection methods can be trained once on background datasets and then deployed universally without per-scene retraining or parameter tuning, showing strong efficiency and robustness. Building upon this paradigm, we propose a decoupled complementary spectral--spatial learning framework for background representation enhancement. The framework follows a two-stage training strategy: (1) we first train a spectral enhancement network via reverse distillation to obtain robust background spectral representations; and (2) we then freeze the spectral branch as a teacher and train a spatial branch as a complementary student (the ``rebellious student'') to capture spatial patterns overlooked by the teacher. Complementary learning is achieved through decorrelation objectives that reduce representational redundancy between the two branches, together with reconstruction regularization to prevent the student from learning irrelevant noise. After training, the framework jointly enhances background representations from both spectral and spatial perspectives, and the resulting enhanced features can be plugged into parameter-free, training-free detectors (e.g., the Reed--Xiaoli (RX) detector) for test-time deployment without per-scene retraining or parameter tuning. Experiments on the HAD100 benchmark demonstrate substantial improvements over representative baselines with modest computational overhead, validating the effectiveness of the proposed complementary learning paradigm. Our code is publicly available at \url{https://github.com/xjpp2016/FERS}.

\end{abstract}


\section{Introduction}
\IEEEPARstart{H}{yperspectral} remote sensing technology acquires data across a broad spectrum of wavelengths from the visible to the infrared regions \cite{ref3,ref4}, delivering rich spectral information. These data have been widely applied in fields such as agriculture \cite{ref6}, resource exploration \cite{ref5}, environmental monitoring \cite{ref7}, and military reconnaissance \cite{ref8}. Among these applications, Anomaly Detection (AD) constitutes a fundamental task in hyperspectral image (HSI) analysis, aiming to identify pixels whose spectral characteristics markedly differ from those of their surrounding background \cite{ref1,ref2}. By highlighting such anomalies, AD facilitates the detection of geological variations, landscape transformations, biodiversity shifts, environmental changes, or human activities—each serving as a critical indicator in its respective domain.

The Reed-Xiaoli (RX) algorithm is a classical baseline for hyperspectral AD \cite{ref9}. It detects anomalous targets by modeling background pixels under a multivariate Gaussian distribution and computes the Mahalanobis distance between each spectral vector and the background model to identify anomalies. However, the original RX algorithm relies solely on global statistical distribution, rendering it susceptible to background non-uniformity and deviations from the Gaussian assumption. To mitigate these limitations, numerous variants have been developed. Local-window RX algorithms (such as LRX \cite{grxlrx}, QLRX \cite{QLRX}, and LAIRX \cite{LAIRX}) utilize localized background statistics to address distribution heterogeneity, while Kernel RX (KRX) \cite{KRX} employs kernel mapping to project data into high-dimensional spaces, thereby handling nonlinear relationships and relaxing the Gaussian constraint. Despite these improvements, such methods generally entail higher computational costs and greater parameter sensitivity compared to the classical RX, and their performance often depends on laborious parameter re-tuning across different testing scenarios.

In recent years, deep learning has introduced new paradigms for hyperspectral AD, with autoencoder-based reconstruction methods becoming particularly prominent. These approaches are generally based on the assumption that anomalies, due to their rarity, cannot be accurately reconstructed by a trained autoencoder. Representative techniques include pixel-wise spectral reconstruction \cite{DBN_HAD,ref16} and full-image reconstruction using 2D convolutional autoencoders \cite{ref10}. However, autoencoders are prone to overfitting and often reconstruct anomalies as effectively as background content, which compromises detection accuracy. The Semi-supervised Background Estimation Model (SBEM) \cite{ref17} mitigates this issue by employing a Generative Adversarial Network (GAN)-based autoencoder \cite{GAN} trained on pre-screened background pixels. Nonetheless, like many deep-learning-based hyperspectral AD methods \cite{survey_DeepHAD}, SBEM relies on a per-image training strategy. This fundamentally limits its generalization capability and incurs substantial computational overhead, as the model requires retraining for each new test scenario.
\IEEEpubidadjcol

To overcome these limitations and enable a train-once, deploy-anywhere HAD framework—requiring no test-time retraining or per-scene parameter tuning—Li et al. proposed the Anomaly Enhanced Transformation Network (AETNet) \cite{ref13}. AETNet injects synthetic anomalies into background HSIs to train a Transformer-based denoising autoencoder. Once trained, it enhances anomalies in unseen HSIs, allowing simple parameter-free detectors (e.g., RX) to achieve strong performance on large-scale benchmarks such as HAD100 \cite{ref13}. However, its effectiveness depends heavily on data augmentation strategies, which may limit scalability.

To address this issue, we previously introduced Feature Enhancement via Reverse Distillation (FERD) \cite{FERD}, whose training requires no data augmentation. Unlike AETNet, which enhances anomaly priors, FERD enhances background representations by combining reverse distillation \cite{RD2022} with a spectral feature alignment mechanism, yielding a lightweight Feature Enhancement Network (FEN) that improves both efficiency and accuracy over AETNet on HAD100. Nevertheless, FERD is primarily tailored to spectral representations, leaving spatial cues underutilized.

\begin{figure}[hbtp]
    \centering
    \includegraphics[width=0.8\linewidth]{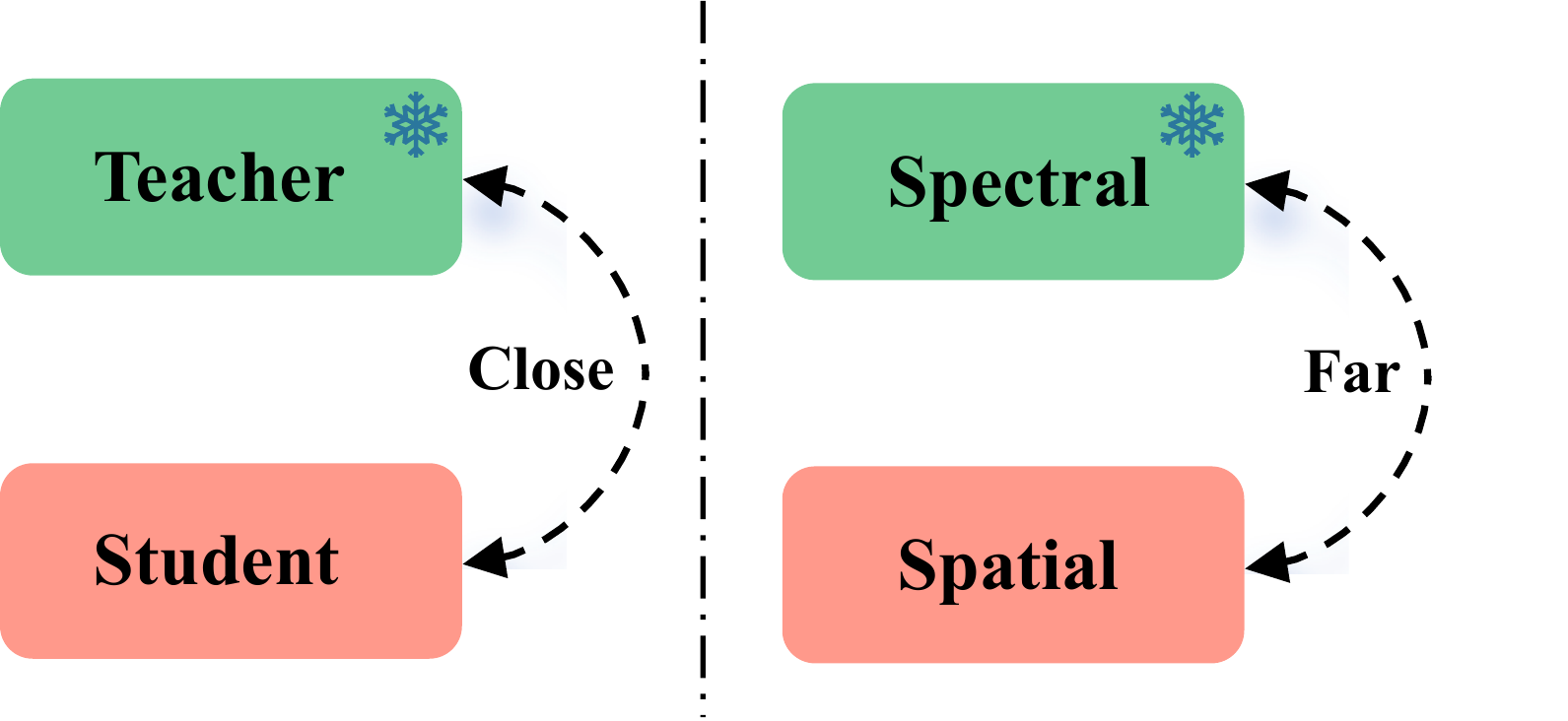}
	\caption{Schematic illustration of the proposed decoupled complementary teacher--student framework. The spectral branch serves as a fixed teacher, while the spatial branch is trained as a complementary student (dubbed the ``rebellious student'') to learn non-redundant cues beyond the teacher, guided by feature decorrelation objectives.}
    \label{fig:G2_Instruction_ST}
\end{figure}

To overcome this limitation, we propose a fully decoupled stepwise training strategy for complementary spectral--spatial background representation enhancement. The strategy proceeds in two stages: we first enhance spectral representations, and then introduce a spatial branch to capture cues that the spectral branch alone may miss.

Specifically, we adopt a teacher--student formulation where the spectral branch serves as a fixed teacher and the spatial branch is trained as a complementary student (the ``rebellious student''). Rather than mimicking the teacher, the student is encouraged to learn non-redundant representations via feature decorrelation objectives (cross-covariance and cosine similarity penalties), together with variance-preserving regularization to avoid collapse. A reconstruction constraint further couples the two branches to recover the input HSI, ensuring that the learned spatial representations are meaningful rather than noise.

\begin{figure}[hbtp]
    \centering
    \includegraphics[width=0.6\linewidth]{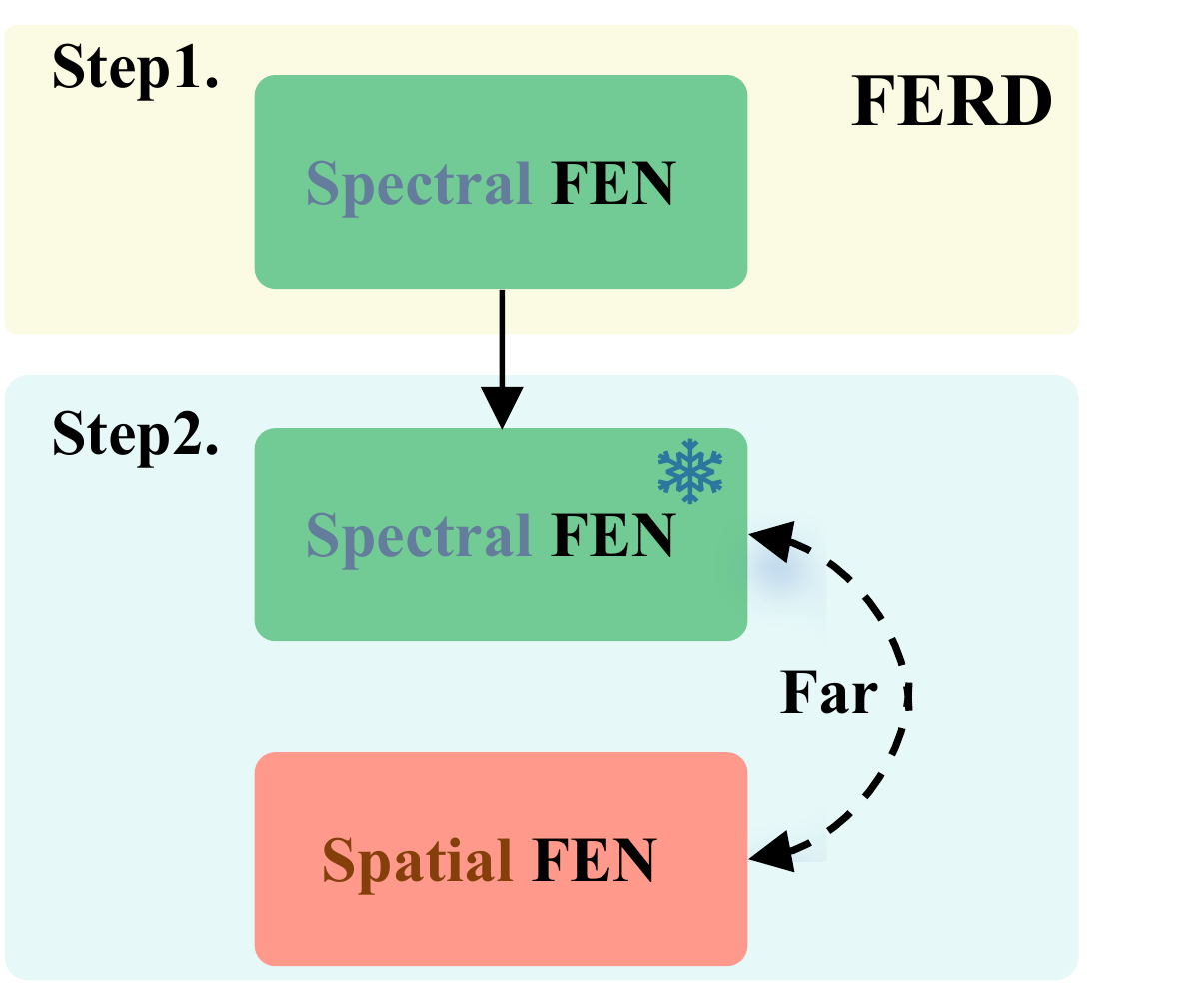}
    \caption{Illustration of the implementation of the rebellious student framework in our method. In Step 1, the spectral Feature Enhancement Network (FEN) is obtained via FERD and used as the teacher network in Step 2. In Step 2, the spatial FEN is trained as the “rebellious student” to capture complementary spatial features.}
    \label{fig:G2_Train_with_2Steps}
\end{figure}

The overall training procedure is summarized in Fig.~\ref{fig:G2_Train_with_2Steps}:
\textbf{Stage 1 --- Spectral Branch Training:} we follow FERD to obtain a spectral FEN specialized for background spectral representation enhancement.
\textbf{Stage 2 --- Spatial Branch Training:} the spectral FEN is frozen as the teacher, and a spatial FEN is trained as the complementary student to learn spatial cues overlooked by the teacher. In addition, we deliberately compress spectral information in the student input to suppress spectral shortcuts and promote spatial learning.

On top of this two-branch design, we further explore two fusion strategies: (1) direct addition of spectral and spatial features at the representation level, and (2) element-wise Hadamard product at the anomaly score level. These strategies ensure robust anomaly localization under different conditions. 

Experimental results on the HAD100 dataset demonstrate that, with only a marginal increase in computation, incorporating spatial information yields substantial improvements over FERD, thereby validating the effectiveness of the proposed complementary learning paradigm. Moreover, cross-scene experiments on other scenes reveal that the learned enhancement mechanism reshapes the spectral features into more Gaussian-like and discriminative distributions, revealing the underlying mechanism through which background feature enhancement benefits both statistical and learning-based detection paradigms.

In summary, the main contributions of this paper are threefold:

\begin{enumerate}
\item We propose a novel \textbf{complementary spectral--spatial learning paradigm} that extends FERD by introducing a teacher--student formulation where a spatial branch is trained as a complementary student (dubbed the ``rebellious student'') to learn \emph{non-redundant} cues beyond the spectral teacher, guided by feature decorrelation objectives.

\item We introduce a \textbf{fully decoupled stepwise training strategy} that first learns a spectral enhancement network via FERD and then freezes it to supervise spatial learning with explicit suppression of spectral shortcuts, enabling progressive and complementary background representation enhancement.

\item We provide \textbf{comprehensive experimental validation and mechanistic insights}: in addition to consistent gains over FERD on HAD100 with modest overhead, cross-scene analyses show that the proposed enhancement regularizes background feature distributions toward more Gaussian-like structures and improves anomaly separability, helping explain its benefits for both statistical and learning-based detectors.
\end{enumerate}

\section{Related Work}

\subsection{Spectral-Spatial Feature Utilization in HSI Anomaly Detection}
In hyperspectral AD, effectively exploiting and balancing spectral and spatial features has always been a central challenge. Existing methods can generally be categorized according to their strategies for spectral-spatial feature fusion.

\textbf{Traditional RX-based methods.} The RX algorithm\cite{ref9} is one of the earliest AD methods, which determines anomalies by calculating the Mahalanobis distance between a test pixel and the background distribution. Variants of RX, such as local-window RX methods (LRX\cite{grxlrx}, QLRX\cite{QLRX}, LAIRX\cite{LAIRX}, etc.), incorporate local spatial information into the RX framework, thereby enhancing the use of spatial features. However, their modeling of spatial features remains relatively shallow.

\textbf{Low-rank and sparse decomposition methods.} Another category of methods relies on the assumption that background components are generally low-rank, while anomalies correspond to sparse perturbations. A typical representative is the application of low-rank and sparse matrix decomposition (LRaSMD)\cite{zhou2011godec} to hyperspectral AD, as in \cite{LRASR2016TGRS}. Such methods are capable of modeling both spectral and spatial dimensions and offer strong theoretical interpretability. However, they usually suffer from high computational complexity and sensitivity to parameter settings.

\textbf{Deep learning methods.} In recent years, deep learning has become a major avenue for achieving deep fusion of spectral-spatial features in hyperspectral AD. Based on how spectral and spatial information is utilized, existing methods can be divided into “integrated” and “branch-based” approaches. Integrated methods typically extract spectral-spatial features in a unified framework to enable mutual complementation. For example, \cite{3D-AEAN,ref10} reconstructs entire images using convolutional autoencoders to jointly capture spectral and spatial features, \cite{3D-CVAE} employs 3D convolutional variational autoencoders (3D-CVAE) for joint spectral-spatial modeling, and \cite{3D-Transformer} introduces a Transformer-based approach that incorporates global self-attention for spectral-spatial feature learning. Similarly, \cite{SSFE} proposes a spatial-spectral features enhancement (SSFE) module that leverages depth-wise separable (DWS) convolution layers\cite{DWS} to strengthen both spectral and spatial representations. By contrast, branch-based methods treat spectral and spatial information as two relatively independent streams, which are fused at a later stage. For instance, \cite{ss-DSVDD} extracts spectral features using 1-D convolutional autoencoders (CAEs) and spatial features using 2-D CAEs, followed by feature concatenation and multilayer perceptron (MLP)-based fusion; \cite{LS3T-Net} adopts an adaptive convolutional network for extracting spatial neighborhood features, while simultaneously employing a variational autoencoder (VAE) to model spectral distributions, and fuses both branches' results at the decision level.

Although integrated methods enable joint modeling of spectral-spatial features, they face two main limitations: (1) spectral and spatial branches may overlap, resulting in redundant modeling, and (2) some critical features may be weakened during integration, thereby impairing anomaly discrimination. Based on this observation, our work explicitly separates spectral and spatial feature extraction within the model structure and enhances complementarity and balance via flexible fusion strategies. Specifically, we adopt a branch-based scheme and employ a stepwise training strategy to fully exploit information from both branches.

\subsection{Negative Correlation Learning}
Although our proposed rebellious student-teacher framework is introduced for the first time in this work, its underlying idea is similar to Negative Correlation Learning (NCL) \cite{Liu1999NCL}, which was initially developed for ensemble learning \cite{dong2020survey, sagi2018ensemble}. Its core idea is to explicitly encourage negatively correlated prediction errors among individual networks by modifying the error function, thereby driving each learner to focus on different aspects of the data, achieving better division of labor, and ultimately improving the generalization of the ensemble. In recent years, NCL has been further applied to deep ensemble learning, with representative cases including CNN ensembles for image classification\cite{buschjaeger2020gncl} and RNN ensembles for time series prediction\cite{peng2020nclrelm}.

Beyond its direct use in ensemble learning, the concept of NCL has been extended to other domains. For example, in noisy-label and robust learning scenarios, \cite{yan2022lacol} proposed leveraging “negative correlation” signals to guide latent contrastive learning by introducing sample-level negative correlation constraints in the metric space, thereby improving classification robustness. In multimodal learning\cite{MML2019,MAVD}, \cite{ICU2024} applied a temporal distance-based soft negative correlation constraint to multimodal ICU electronic health records (EHR, including clinical notes and time-series data) for online prediction tasks. This effectively mitigated the shortcomings of traditional contrastive learning in modeling gradual temporal state changes in healthcare, leading to significant improvements in zero-shot deterioration prediction. Intuitively, whenever complementarity among different branches (learners, tasks, modalities) is desired, incorporating NCL often boosts performance. However, to the best of our knowledge, relevant research remains relatively limited.

In summary, NCL differs from our proposed method in key aspects. While both approaches aim to encourage complementary learning, our goal is to train an entirely new network rather than simply enhancing existing ensemble or collaborative learning frameworks.

\section{Proposed Method}
This section presents a decoupled complementary spectral--spatial teacher--student paradigm for HSI background prior modeling, which boosts the performance of parameter-free, test-time training-free detectors (e.g., RX) without per-scene retraining or parameter tuning.

In this paradigm, the Spectral FEN (Spe-FEN) is designated as the teacher network, while the Spatial FEN (Spa-FEN) acts as the “rebellious student.” Unlike conventional teacher-student frameworks, the student network in our method does not mimic the teacher's behavior; instead, it focuses on learning complementary spatial features overlooked by the teacher. This leads to functional synergy and complementarity between the two networks.  

The overall training process is divided into two stages:  

\textbf{Stage 1: Teacher Network Training.} The spectral branch (Spe-FEN) is fully trained to enhance its capability of modeling background spectral characteristics. After training, the parameters of this branch are frozen and serve as the teacher model for the subsequent student network training.  

\textbf{Stage 2: Student Network Training.} Guided by the teacher, the spatial branch (Spa-FEN) focuses on extracting complementary spatial information that the spectral branch fails to capture, thereby enhancing the overall background prior modeling ability.  

All training procedures are conducted on background datasets, with the detailed process described below.  

\subsection{Stage 1: Training the Teacher Network (Spectral Branch)}

\begin{figure*}[t]
    \centering
    \includegraphics[width=0.95\linewidth]{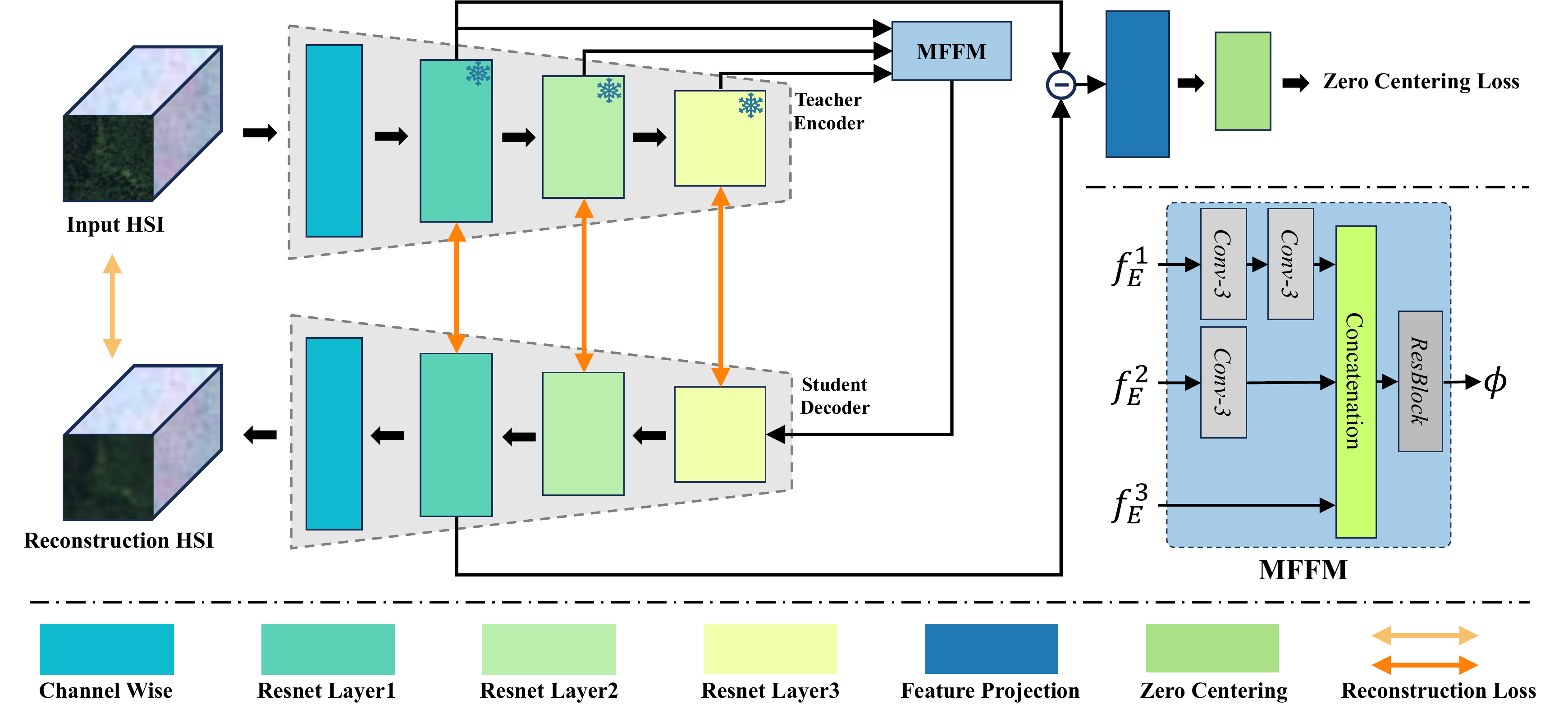}
    \caption{An illustration of the encoder-decoder with reverse distillation and the Spectral Feature Alignment Mechanism (SFAM) in \textbf{Stage 1 (Spectral Branch)}. The reverse distillation differs from a standard encoder-decoder as most of the encoder is frozen, analogous to the teacher in knowledge distillation. However, it diverges from standard distillation by having a reversed data flow in the student (decoder). The SFAM, primarily realized through the Zero Centering Loss and its auxiliary structures, aims to softly align the most spectrally-relevant features from the encoder and decoder in the spectral space.}
    \label{fig:ferd_pipeline}
\end{figure*}

In the first stage, the spectral feature enhancement branch is trained as the teacher network. This branch employs a reverse distillation mechanism and a spectral feature alignment mechanism (SFAM) to strengthen spectral representations, while structural pruning preserves only the most relevant components for spectral modeling. The details of this process are elaborated in the following three subsections.

\textbf{1) Encoder-decoder with reverse distillation.}  
The term ``reverse distillation'' stems from its data flow direction: unlike conventional distillation where both teacher and student follow the same forward path, here they operate in opposite directions, resembling an autoencoder in structure. In this setup, the teacher (encoder) maps HSIs into feature representations, while the student (decoder) reconstructs the HSIs from these features. However, this is not a typical autoencoder, as most of the encoder's structure remains fixed. This results in an asymmetric architecture that helps mitigate overfitting.

The encoder is built upon a ResNet backbone, with a Channel-Wise Layer (CWL) prepended to adapt to the spectral dimension of HSIs. Given an input HSI $H$, the encoder extracts multi-scale spectral features $f_{E}^{1}, f_{E}^{2}, f_{E}^{3}$. Except for the CWL, the encoder parameters are frozen, and FERD only updates the CWL during training to better capture basic spectral features.  

The multi-scale feature fusion module (MFFM) then fuses hierarchical encoder features into a compact representation $\phi$, which is fed into the decoder. The decoder is structurally symmetric to the encoder but uses transposed convolutions instead of convolutions, producing reconstruction features $g_{D}^{3}, g_{D}^{2}, g_{D}^{1}$. Finally, the decoder's CWL maps these features back into the spectral domain to yield the reconstructed HSI.  

FERD optimizes the reverse distillation process using a combination of feature similarity loss and reconstruction loss:
\begin{equation}
\mathcal{L}_{\text{sim}} = 1 - \frac{1}{N} \sum_{i=1}^{N} \text{CosSim}(g_{D}^i, f_{E}^i), 
\quad
\mathcal{L}_{\text{mse}} = \| H - H_R \|_2^2,
\end{equation}
where $H$ is the input HSI, $H_R$ is its reconstruction, and $\text{CosSim}(\cdot, \cdot)$ denotes the cosine similarity function. The feature similarity loss $\mathcal{L}_{\text{sim}}$ promotes alignment between the multi-scale features of the encoder ($f_E^i$) and decoder ($g_D^i$), while the pixel-wise reconstruction loss $\mathcal{L}_{\text{mse}}$ ensures the output faithfully reproduces the input HSI.

\textbf{2) Spectral Feature Alignment Mechanism (SFAM).}  
To capture advanced spectral representations without introducing extraneous noise, we introduce the SFAM. SFAM is designed to align low-level features from the encoder ($f_{E}^{1}$) with their corresponding features from the decoder ($g_{D}^{1}$), which have acquired higher-level characteristics after passing through advanced structures in the network. Specifically, SFAM comprises a Feature Projection Layer (FPL) and a Zero-Centralization Layer (ZCL). The FPL projects both $f_{E}^{1}$ and $g_{D}^{1}$ into a shared spectral space, obtaining $f_{E}^{P}$ and $g_{D}^{P}$. Since $g_{D}^{P}$ may contain noise artifacts introduced by the complex model transformations it has undergone, we employ a soft alignment strategy: we compute the absolute difference $|f_{E}^{P} - g_{D}^{P}|$, project it into a one-channel representation $O$, and apply zero-mean regularization:
\begin{equation}
\mathcal{L}_{\text{Z}}(O) = -\frac{1}{N}\sum_{i=1}^{N} \log\left(1 - \frac{1}{1+e^{-o_i}}\right).
\end{equation}  

The overall training loss integrates all terms:  
\begin{equation}
\mathcal{L} =  \mathcal{L}_{\text{sim}}  + \lambda_{\text{mse}} \mathcal{L}_{\text{mse}} + \lambda_{\text{Z}} \mathcal{L}_{\text{Z}}.
\end{equation}

\begin{figure}[t]
    \centering
    \includegraphics[width=0.9\linewidth]{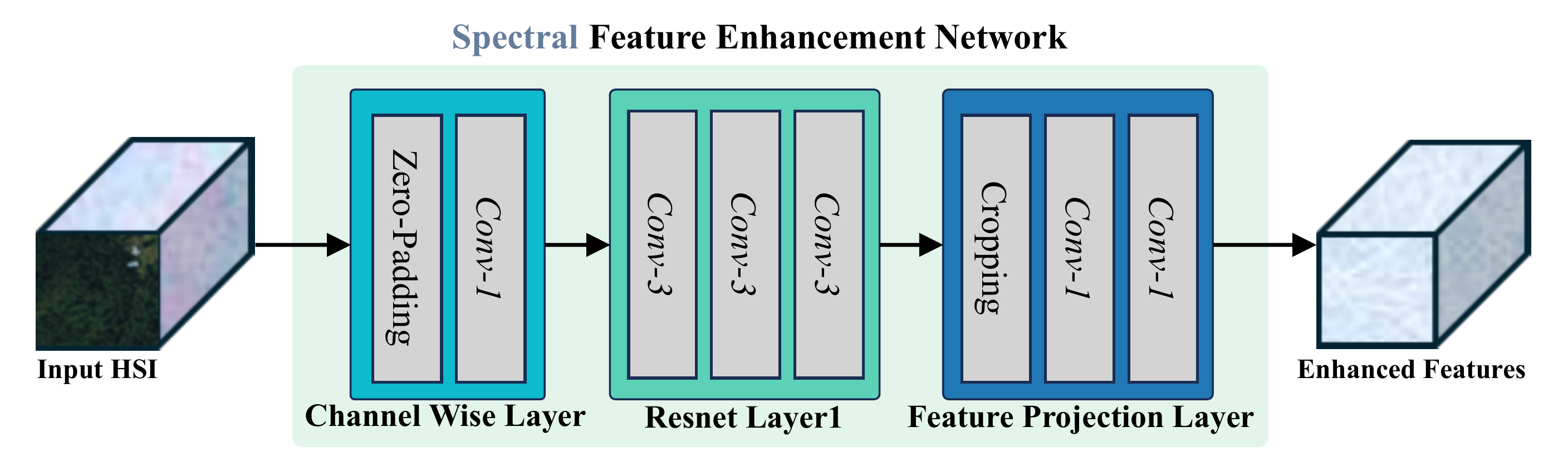}
    \caption{The pruned Spectral Feature Enhancement Network (Spe-FEN), which retains only the components most relevant to spectral feature extraction—the shallow layers of the encoder and the Feature Projection Layer (FPL) from the SFAM. Its architecture predominantly comprises 1×1 convolutions, with only a few 3×3 convolutions, ensuring its focus on modeling spectral characteristics.}
    \label{fig:spe_fen}
\end{figure}

\textbf{3) Pruning to obtain the Spectral Feature Enhancement Network (Spe-FEN).}  
To improve efficiency and reduce redundancy, we retain only the components most relevant to spectral feature extraction—namely, the CWL layer, ResNet layer1, and the FPL in SFAM—and define them together as the Spe-FEN. As shown in Fig.~\ref{fig:spe_fen}, the input HSI is processed by Spe-FEN to produce spectral features $F_{spe}$. Except for the three $3\times 3$ convolutions inside each ResBlock, all other layers adopt $1\times 1$ convolutions, ensuring $F_{spe}$ focuses on spectral features:  
\begin{equation}
F_{spe} = \text{FEN}_{spe}(H).
\end{equation}
This network is later employed as the teacher model in Stage 2.

\subsection{Stage 2: Training the Student Network (Spatial Branch)}  

With the teacher network trained and fixed, the second stage optimizes the spatial feature enhancement branch. This branch is designed as a ``rebellious student,'' diverging from the typical mimicry paradigm by focusing on the spatial features complementary to the teacher's spectral expertise, thus boosting overall detection performance.

The Spa-FEN architecture is built for this purpose with three core components: spectral compression to minimize interference, a ResNet for high-level spatial feature extraction, and a module with multi-scale pooling and SEBlock for spatial enhancement. To guide the learning, cross-covariance and cosine similarity constraints decorrelate student-teacher features, supported by variance-preserving regularization to avoid collapse. Finally, a lightweight decoder (stacked channel attention and residual blocks) fuses the features and reconstructs the HSI, ensuring the student learns meaningful spatial complements rather than redundant noise.

The three core components of Spa-FEN are detailed below:

\textbf{(1) Spectral Compression and Spatial Focus.}  
The spectral compression module projects the input HSI with tens or hundreds of bands into a three-channel representation $H_{3c}$. This process is formulated as:
\begin{equation}
H_{3c} = \sigma(\, W_2 \, * \, \text{ReLU}(\, W_1 \, * \, H \,) \,),
\end{equation}
where $H$ is the input HSI cube, $*$ denotes the $1\times1$ convolution, and $\sigma$ is the Sigmoid function. This intentional bottleneck forces the network to discard spectral redundancy and prioritize spatial structures.

\textbf{(2) Deep Spatial Feature Extraction.}  
The compressed representation $H_{3c}$ is fed into a frozen ResNet-34 backbone (pretrained on ImageNet) to extract deep 256-channel spatial features. Specifically, the network passes through the initial convolution, max pooling, and the first three residual blocks (\texttt{layer1}-\texttt{layer3}). The ResNet parameters remain fixed to prevent overfitting and to leverage general-purpose spatial representations learned from natural images.

\textbf{(3) Multi-scale Context Enhancement.}  
We introduce a Lightweight Spatial Pyramid Pooling module (LightSPP) to capture multi-scale contextual information. For an input feature map $f_{\text{res}}$(from ResNet-34), it applies adaptive average pooling at scales $\{1,2,4\}$, upsamples all outputs to the original spatial dimensions, and fuses them:
\begin{equation}
\begin{aligned}
f_{\text{avp}}^s &= \text{Resize}(\text{AvgPool}_s(f_{\text{res}})), \quad s \in \{1, 2, 4\}, \\
f_{\text{cat}} &= [f_{\text{avp}}^1, f_{\text{avp}}^2, f_{\text{avp}}^4], \\
f_{\text{fuse}} &= W_{\text{fuse}} * \ f_{\text{cat}},
\end{aligned}
\end{equation}
where $[\cdot]$ denotes channel-wise concatenation and $W_{\text{fuse}}$ is a $1\times1$ convolution for feature fusion. Finally, a Squeeze-and-Excitation (SE) Block is applied for channel-wise recalibration:
\begin{equation}
F_{spa} = \text{SE-Block}(f_{\text{fuse}}),
\end{equation}
enhancing informative spatial channels and suppressing less useful ones.

Together, these components form the Spa-FEN, yielding the spatial feature representation:
\begin{equation}
F_{spa} = \text{FEN}_{spa}(H).
\end{equation}

To enable complementary interaction with the spectral teacher, the spatial features $F_{spa}$ are projected into the same dimensional space as the spectral features $F_{spe}$. Their fused representation is then fed into the decoder for HSI reconstruction, ensuring a comprehensive integration of both feature types.

\begin{figure*}[t]
    \centering
    \includegraphics[width=0.95\linewidth]{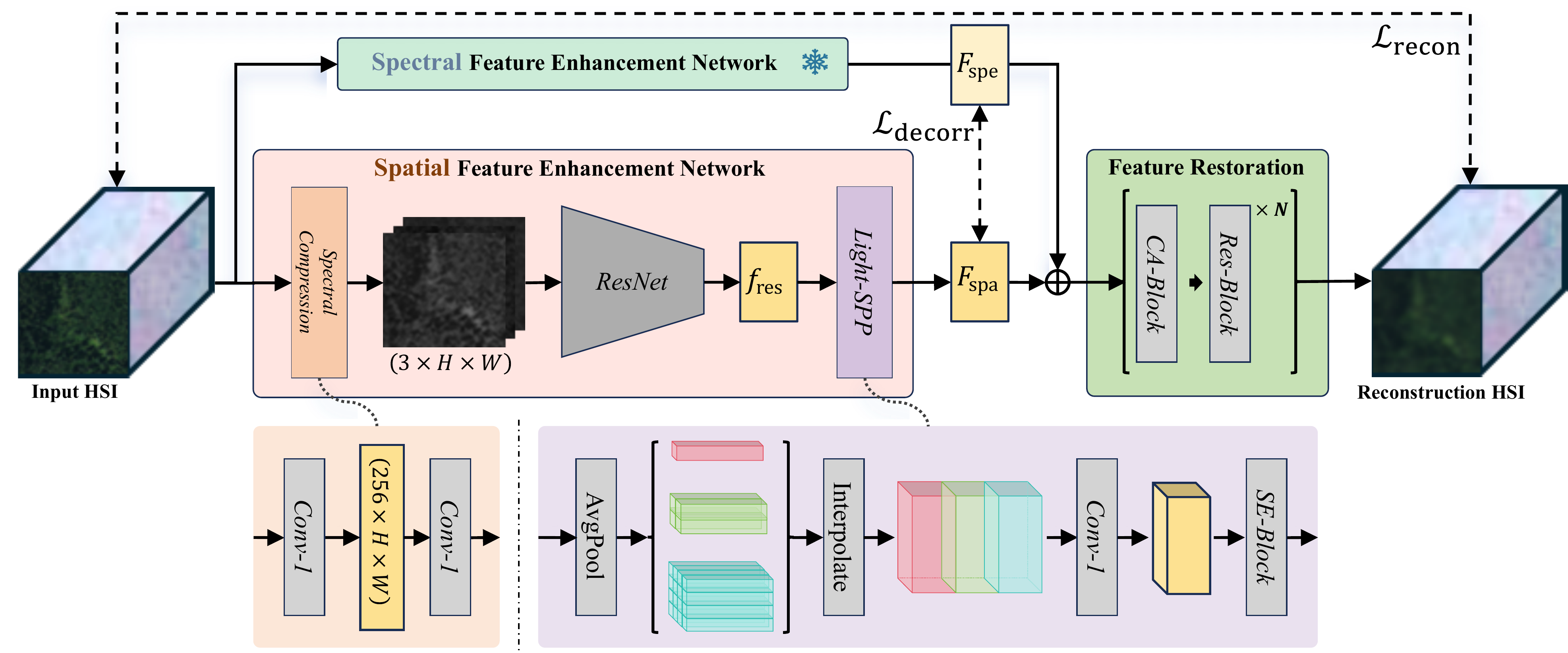}
    \caption{Architecture and training process of the \textbf{Spatial Branch (Stage 2)}. During this stage, the pre-trained Spectral Feature Enhancement Network (Spe-FEN) serves as a fixed teacher, while the Spatial Feature Enhancement Network (Spa-FEN) acts as a ``rebellious student'' to learn spatial features complementary to the teacher's spectral expertise. The Spa-FEN comprises: a spectral compression module to reduce interference, a ResNet-34 for high-level spatial feature extraction, and a LightSPP module for multi-scale context aggregation. A decoder composed of stacked channel attention and residual convolutional layers then fuses the features. The training objective combines a decorrelation loss $\mathcal{L}_1$ to encourage feature independence between the teacher and student, and a reconstruction loss $\mathcal{L}_2$ to ensure the student outputs meaningful features for HSI reconstruction.}
    \label{fig:spa_train}
\end{figure*}

\textbf{Feature Restoration Decoder.}  
We design a lightweight Feature Restoration Network (FRN), which takes the fused features $F_{spe} + F_{spa}$ and reconstructs the HSI:  
\begin{equation}
\hat{H} = \text{FRN}(F_{spe} + F_{spa}).
\end{equation}  
Unlike conventional decoders with upsampling or transposed convolutions, FRN maintains resolution throughout, relying solely on residual and attention modules for feature calibration. This ensures pixel-level correspondence and avoids unnecessary interpolation errors.  

\textbf{Training Objectives.}
The spatial branch (student) is trained with a composite loss function composed of a feature decorrelation loss \(\mathcal{L}_{\text{decorr}}\) and a hyperspectral image (HSI) reconstruction loss \(\mathcal{L}_{\text{recon}}\):
\begin{equation}
\mathcal{L} = \mathcal{L}_{\text{decorr}} + \lambda_{\text{recon}} \cdot \mathcal{L}_{\text{recon}}.
\end{equation}

\textbf{Feature Decorrelation Loss.} This composite loss ensures that the student network learns spatial features that are both statistically independent and directionally orthogonal to the teacher's spectral features, thereby capturing truly complementary information. It integrates three complementary objectives:
\begin{equation}
\mathcal{L}_{\text{decorr}} = \mathcal{L}_{\text{cc}} + \lambda_{\text{cos}} \cdot \mathcal{L}_{\text{cos}} + \lambda_{\text{var}} \cdot \mathcal{L}_{\text{var}}.
\end{equation}

\begin{itemize}
    \item \textit{Cross-Covariance Loss (\(\mathcal{L}_{\text{cc}}\))}: This loss minimizes the linear dependence between the spectral features \(F_{spe}\) and spatial features \(F_{spa}\). Let \(X \in \mathbb{R}^{N \times D_s \times (H \cdot W)}\) and \(Y \in \mathbb{R}^{N \times D_t \times (H \cdot W)}\) be the whitened (zero-mean and unit-variance) and flattened feature maps from the teacher and student, respectively. The cross-covariance matrix is computed as:
    \begin{equation}
    \mathbf{C} = \frac{1}{N \cdot L} X Y^\top,
    \end{equation}
    where \(L = H \cdot W\) is the number of spatial locations. The loss is defined as the squared Frobenius norm of the cross-covariance matrix:
    \begin{equation}
    \mathcal{L}_{\text{cc}} = \|\mathbf{C}\|_F^2 = \sum_{i=1}^{D_s} \sum_{j=1}^{D_t} |C_{ij}|^2.
    \end{equation}
    By minimizing this loss, we encourage the features from the two networks to be statistically decorrelated, eliminating linear dependencies.
    
    \item \textit{Cosine Similarity Loss (\(\mathcal{L}_{\text{cos}}\))}: This loss complements the cross-covariance loss by enforcing \emph{directional orthogonality} between the spectral and spatial features. While the cross-covariance loss minimizes linear correlations, the cosine similarity loss specifically encourages the feature vectors to be perpendicular or opposite in direction:
    \begin{equation}
    \mathcal{L}_{\text{cos}} = 1 + \text{CosSim}\left(F_{spe}, F_{spa}\right),
    \end{equation}
    where $\text{CosSim}(\cdot, \cdot)$ is the cosine similarity function. The loss reaches its minimum value of 0 when the features are discouraging directional alignment (encouraging complementary directions) (\(\text{CosSim} = -1\)), and its maximum of 2 when they are perfectly aligned (\(\text{CosSim} = +1\)). By penalizing directional alignment, this loss ensures that the spatial features capture information in directions orthogonal to the spectral features.
    
	\item \textit{Variance-Preserving Loss (\(\mathcal{L}_{\text{var}}\))}:  
	This loss aims to preserve informative and diverse spatial representations by preventing feature collapse while gently encouraging sparsity.  
	It consists of two complementary components:
	\begin{equation}
	\mathcal{L}_{\text{var}} = \frac{1}{D} \sum_{k=1}^{D} \max\left(0, \tau - \text{Var}(F_{spa}^{(k)})\right) + \| F_{spa} \|_1,
	\end{equation}
	where \(D\) denotes the channel dimension of the spatial features, \(\tau = 1.0\) is the variance threshold, and \(\| \cdot \|_1\) represents the L1 norm.  
	The first term preserves adequate activation variance within each channel, ensuring that spatial features remain informative and non-collapsed.  
	The L1 term serves as a mild regularizer to suppress redundant responses, allowing the network to highlight only the most relevant spatial activations.  
	This design emphasizes maintaining useful diversity while promoting compact and discriminative spatial representations in the enhanced spectral feature space.
\end{itemize}

\textbf{Reconstruction Loss.} This loss serves as an anchoring objective that ensures the student's spatially-enhanced features remain informative for the primary HSI reconstruction task. By requiring accurate reconstruction of the input HSI, it prevents the network from learning arbitrary noise under the decorrelation constraints and guarantees that the captured spatial features represent meaningful visual patterns. The loss combines pixel-level accuracy and structural preservation:
\begin{equation}
\mathcal{L}_{\text{recon}} = \| H - \hat{H} \|_2^2 + \lambda_{\text{ssim}} \cdot \mathcal{L}_{\text{SSIM}}(H, \hat{H}),
\end{equation}
where \(\mathcal{L}_{\text{SSIM}} = 1 - \text{SSIM}(H, \hat{H})\), \(H\) is the input HSI, and \(\hat{H}\) is the reconstructed output. The MSE term enforces pixel-wise fidelity, while the SSIM term maintains structural consistency, together ensuring that the learned complementary features are semantically relevant rather than random noise.

\subsection{Inference and Anomaly Scoring}  
\label{method_fusion}

\begin{figure}[t]
    \centering
    \includegraphics[width=0.9\linewidth]{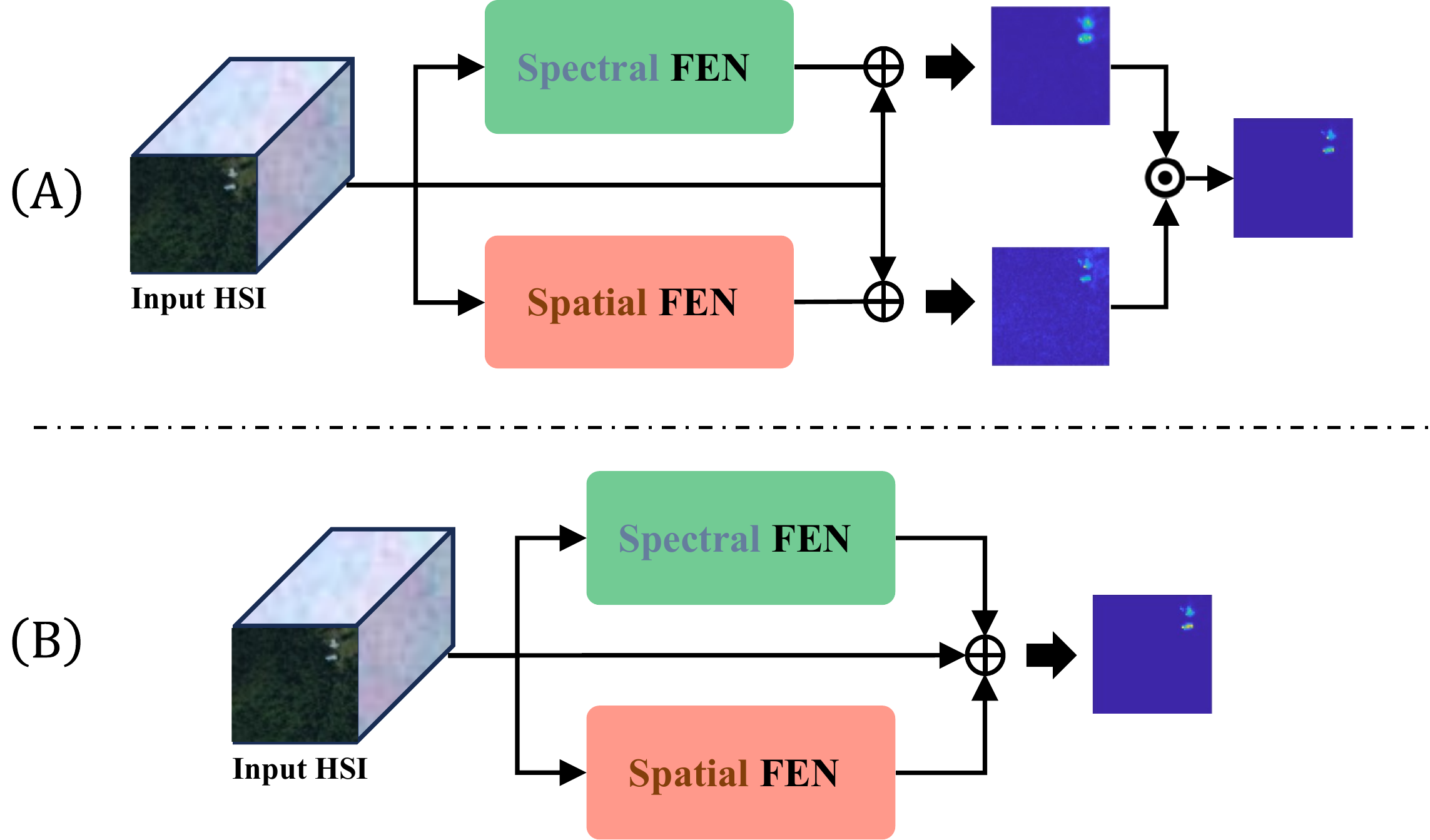}
    \caption{A schematic overview of the proposed fusion strategies. (A) Score-level fusion (Multiplicative): Independent RX detection on each branch, followed by score normalization and multiplication. (B) Feature-level fusion (Additive): Early integration of spectral and spatial features prior to a single RX detection process.}
    \label{fig:fusion_AD}
\end{figure}

Anomaly detection at inference time builds upon the RX algorithm, which computes the Mahalanobis distance of each pixel relative to the background statistics ($\boldsymbol{\mu}$, $\mathbf{\Sigma}$):
\begin{equation}
    S_{\text{RX}}(i) = (h_E^i - \boldsymbol{\mu})^\top \mathbf{\Sigma}^{-1} (h_E^i - \boldsymbol{\mu}).
\end{equation}

We introduce two fusion strategies to combine spectral ($F_{spe}$) and spatial ($F_{spa}$) features for enhanced anomaly detection, as visualized in Fig.~\ref{fig:fusion_AD}:

\paragraph{Normalized Multiplicative Fusion.}
This strategy performs score-level fusion by computing RX scores independently for each branch, followed by min-max normalization and element-wise multiplication:
\begin{align}
    S_{spa} &= \text{RX}(H + F_{spa}), \\
	\quad S_{spe} &= \text{RX}(H + F_{spe}), \\
    \widetilde{S}_{spa} &= \frac{S_{spa} - \min(S_{spa})}{\max(S_{spa}) - \min(S_{spa})}, \quad  \\
    \widetilde{S}_{spe} &= \frac{S_{spe} - \min(S_{spe})}{\max(S_{spe}) - \min(S_{spe})}, \\
    S_{\text{mult}} &= \widetilde{S}_{spa} \odot \widetilde{S}_{spe}.
\end{align}
The multiplicative fusion essentially acts as a logical AND operation, emphasizing only the regions unanimously identified by both modalities, which effectively suppresses false alarms \cite{ISSTAD}. This approach offers a more direct utilization of spatial information and is particularly effective in scenarios where spatial cues are dominant or the spectral dimensionality is relatively low.

\paragraph{Additive Feature Fusion.}
This strategy performs early fusion by combining features in the spectral domain prior to detection:
\begin{equation}
    S_{\text{add}} = \text{RX}(H + F_{spa} + F_{spe}).
\end{equation}
This method integrates the spatial features $F_{spa}$ into the spectral space, where they complement the spectrally enhanced features $F_{spe}$. By doing so, it enriches the spectral representation with complementary spatial contexts, creating a more informative and discriminative feature set for the detector. This approach is particularly meaningful in spectrally dominant scenarios with high dimensionality, as it enhances the spectral space without altering the fundamental detection mechanism.

\section{Experiments}
\subsection{Datasets and Evaluation Metrics}

\textbf{Training Background Scenarios.} 
Our model is trained on the background scenes from the HAD100 dataset~\cite{ref13}, which offers two distinct training sets captured by AVIRIS-NG and AVIRIS-Classic sensors, respectively. These sets contain exclusively background pixels sampled from diverse natural environments such as grasslands, forests, farmlands, deserts, lakes, rivers, and coastlines. All HSIs are uniformly processed into 64$\times$64 pixel blocks to ensure consistency. 

The selection of training data is guided by the imaging sensor used in the test scenarios to ensure sensor consistency. Accordingly, the AVIRIS-NG training subset is used for evaluation on the HAD100 test set, as they share the same sensor. For experiments on the ABU dataset, which was captured by the AVIRIS-Classic sensor, we employ the corresponding AVIRIS-Classic training set to maintain this sensor alignment.

\textbf{Test Scenarios.} 
Evaluation is primarily conducted on the comprehensive HAD100 test benchmark~\cite{ref13}, comprising \textbf{100} authentic remote sensing scenes with meticulously annotated ground truth. All test HSIs were acquired by AVIRIS-NG and preprocessed into 64$\times$64 pixel patches. The dataset encompasses a wide variety of background types and contains anomalies that are predominantly compact man-made objects, such as vehicles, boats, and buildings. To the best of our knowledge, HAD100 is uniquely positioned as the only large-scale benchmark for evaluating how well methods can leverage background priors, owing to its provision of a dedicated, large-scale background HSI dataset (as described previously) that is free of targets and aligned with the test scenes. The substantial scale and diversity of HAD100 present significant challenges for model generalization—a critical requirement in practical hyperspectral anomaly detection applications.

To further assess the generalizability and background enhancement capability of our method, we also evaluate it on additional scenes captured by AVIRIS sensors. Although our model is trained on 64$\times$64 patches, it imposes no constraint on the input HSI size during inference, allowing evaluation on test scenes with varying resolutions. We select six widely-used scenes from the Airport-Beach-Urban (ABU) dataset: Airport-1, Airport-2, Airport-3, Airport-4, Beach-4, and Urban-3 (the latter two denoted simply as ``Beach'' and ``Urban''). These scenes vary in spatial and spectral dimensions: Airport-1, -2, and -3 have 100$\times$100 pixels and 205 bands; Airport-4 and Urban have 100$\times$100 pixels and 191 bands; and Beach has 150$\times$150 pixels and 102 bands. These scenes are chosen for their diversity in both background and target characteristics, offering a rigorous test of our method's generalization ability. Notably, several of these scenes are known to be challenging for the standard RX detector, thereby providing a meaningful basis for demonstrating the effectiveness of our proposed background feature enhancement.

\textbf{Evaluation Metrics.}
We employ the Area Under the Receiver Operating Characteristic Curve (AUC) as the primary quantitative evaluation metric, as it is the most widely adopted measure in hyperspectral anomaly detection. The ROC curve characterizes the trade-off between the detection probability ($P_d$) and the false alarm rate ($P_f$) across all decision thresholds. The AUC integrates this information into a single scalar value, where a higher score indicates better detection performance, and a value of 0.5 corresponds to random guessing.

Given the substantial size of the HAD100 benchmark (100 test scenes), we report the mean AUC (mAUC) across all scenes to provide a comprehensive and concise summary of overall detection performance. Presenting individual AUCs for each scene would be impractical. For the smaller ABU dataset, we report the per-image AUC to facilitate a more detailed, scene-wise analysis.

\subsection{Experimental Settings}

\textbf{Comparative Methods on HAD100}
We evaluate our method against several relevant approaches on the HAD100 benchmark. The compared methods include:
\begin{itemize}
    \item \textbf{RX} \cite{ref9}: The classical Global Reed-Xiaoli detector, which also serves as the foundational detector in our proposed framework.
    \item \textbf{Auto-AD} \cite{ref10}: An autoencoder-based method that utilizes 2D convolutions to jointly model spectral and spatial information for reconstruction.
    \item \textbf{LREN} \cite{LREN}: A method employing a trainable Low-Rank Embedded Network to exploit spectral-spatial characteristics.
    \item \textbf{RD4AD} \cite{RD2022}: The original Reverse Distillation framework for anomaly detection. To adapt it for HSI, we replaced the first layer of its encoder and the last layer of its decoder with a CWL, while keeping the rest of the architecture unchanged from the official implementation.
    \item \textbf{AETNet} \cite{ref13}: A test-time anomaly detection method, pioneeringly developed on HAD100, which enhances anomaly priors via data augmentation.
    \item \textbf{FERD} \cite{FERD}: Our prior work, a background spectral feature enhancement method based on reverse distillation.
\end{itemize}
For RX, a widely-used unofficial implementation is employed, while all other comparative methods use their official code. Furthermore, to better validate the effectiveness of autoencoder-based methods, we designed a lightweight autoencoder baseline (denoted as \textbf{AE}) for comparison, where both the encoder and decoder consist of three $1\times1$ convolutional layers. This method identifies anomalies based on HSI reconstruction error and is included in our code repository.

\textbf{Comparative Methods on ABU}
The experiments on the ABU dataset primarily aim to validate the generalizability and background feature enhancement capability of our method. We report the performance of the baseline RX detector, the AE, and Auto-AD, both with and without being augmented by our feature enhancement network.

\textbf{Parameter Settings}
The training configuration is consistent across both stages of our method. We use the Adam optimizer with a learning rate of 0.005, momentum parameters $\beta_1=0.9$ and $\beta_2=0.95$, and a batch size of 16. Training runs for 60 epochs.

The loss weight coefficients are set as follows:
\begin{itemize}
    \item \textbf{Stage 1 (Spectral Branch)}: $\lambda_{\text{mse}} = 0.1$, $\lambda_{\text{Z}} = 0.1$.
    \item \textbf{Stage 2 (Spatial Branch)}: $\lambda_{\text{recon}} = 1.0$, $\lambda_{\text{cos}} = 0.1$, $\lambda_{\text{var}} = 0.1$, $\lambda_{\text{ssim}} = 0.01$.
\end{itemize}

All experiments were conducted on a system with an Intel i7-11700T CPU, 64 GB RAM, and an NVIDIA GeForce RTX 3090 Ti GPU. The code is implemented in Python 3.10 and PyTorch 2.7, with CUDA version 12.8.

\subsection{Experimental Results On HAD100 Dataset}

\begin{table*}[htbp]
	\centering
	\caption{Performance comparison of methods on HAD100}
	\renewcommand{\arraystretch}{1.1} 
	\begin{adjustbox}{max width=1\linewidth}
		\begin{tabular}{c *{6}{c}}
			\toprule
			\multirow{2}{*}{Method} & \multicolumn{2}{c}{The First 50 Bands} & \multicolumn{2}{c}{The First 100 Bands} & \multicolumn{2}{c}{The First 200 Bands} \\
			\cmidrule(lr){2-3} \cmidrule(lr){4-5} \cmidrule(lr){6-7}
			& \multicolumn{1}{c}{\textbf{mAUC}} & \multicolumn{1}{c}{\textbf{Time}} & \multicolumn{1}{c}{\textbf{mAUC}} & \multicolumn{1}{c}{\textbf{Time}} & \multicolumn{1}{c}{\textbf{mAUC}} & \multicolumn{1}{c}{\textbf{Time}} \\
			\midrule
			RX   & 0.9799 & \bfseries 0.017 & 0.9714 & \bfseries 0.024 & 0.9649 & 0.052 \\
			AE & 0.9339 &  0.851 & 0.9096 & 0.862 & 0.9170 & 0.875 \\
			Auto-AD & 0.8636 & 7.236 & 0.7040 & 7.412 & 0.6825 & 7.527 \\
			LREN & 0.8858 &  33.437 & 0.8820 & 35.389 & 0.8771 & 38.523 \\
			RD4AD & 0.8053 & 0.027 & 0.7542 & 0.027 & 0.8196 & \bfseries 0.028 \\
			AETNet & 0.9925 & 0.043 & 0.9875 & 0.049 & 0.9818 & 0.078 \\
			\midrule
			\rowcolor{lightgray}
			FERD  &  0.9941 & 0.037 &  0.9901 & 0.044 &  0.9891 & 0.073 \\
			\rowcolor{darkgray}
			Proposed Method & \bfseries 0.9953 & 0.071 & \bfseries 0.9910 & 0.046 & \bfseries 0.9902 & 0.075 \\
			\bottomrule
		\end{tabular}%
	\end{adjustbox}
	\label{tab:had100_result}%
\end{table*}%

\begin{figure*}[!htbp]
	\centering
	\includegraphics[scale=0.16]{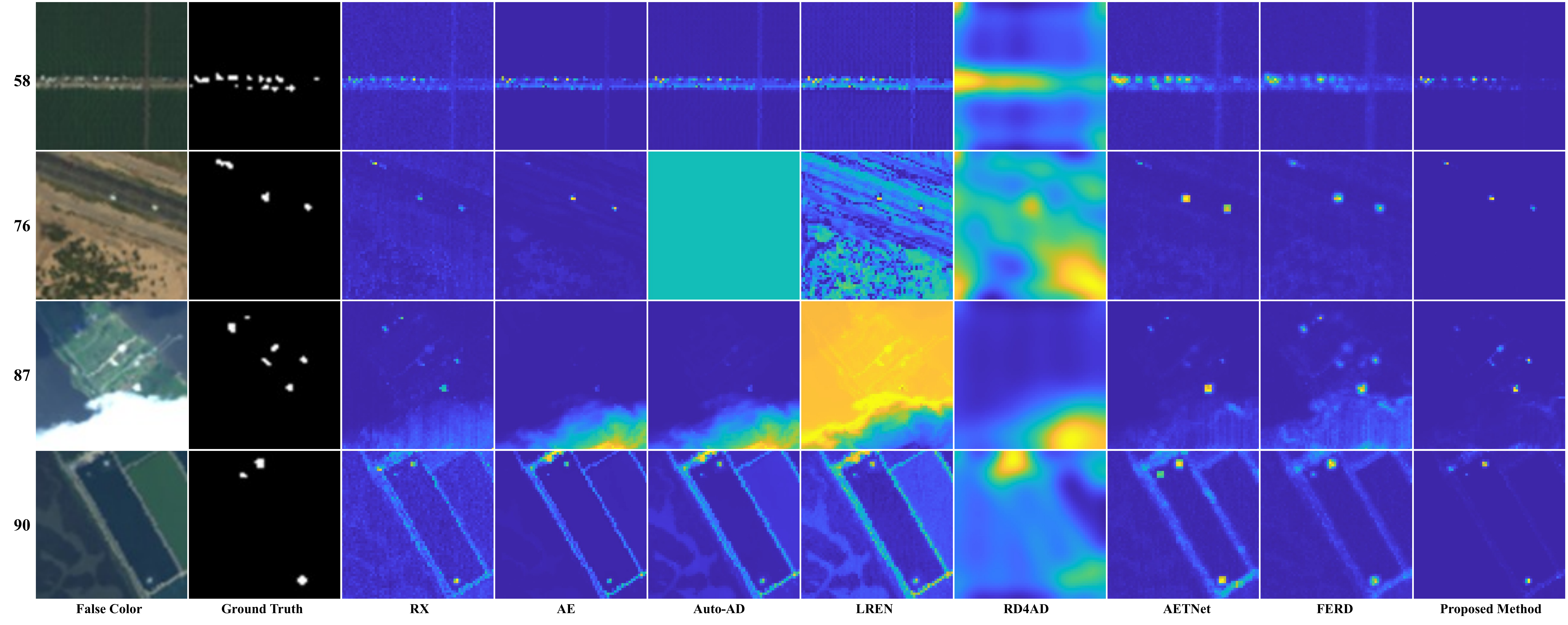} 
	\caption{Illustrative detection maps of different AD methods on the first 50 bands, showcasing four representative examples (Scenes 58, 76, 87, 90) from the 100 test scenes in the HAD100 dataset.}
	\label{scores}
\end{figure*}

In Table \ref{tab:had100_result}, we present a comprehensive performance comparison of various anomaly detection methods on the HAD100 test dataset, evaluated in terms of mean AUC (mAUC) and average computation time (in seconds). The results reveal several key observations.

Methods based on learnable networks that require per-scene training and parameter tuning, such as Auto-AD and LREN, exhibit notably long inference times and struggle to maintain robust performance across large-scale test scenes, despite extensive manual hyperparameter optimization. In contrast, the lightweight autoencoder (AE) achieves relatively stable results, which can be attributed to its simple architecture and minimal parameter sensitivity—in practice, only the learning rate significantly influences its performance. The classical RX detector also demonstrates consistent and stable performance across different band settings.

The original reverse distillation method (RD4AD), while effective in conventional image anomaly detection, shows limited efficacy when directly applied to hyperspectral imagery. On the other hand, AETNet—a parameter-free approach that enhances anomaly priors without per-image retraining—achieves strong performance with low computational cost.

Our previously proposed background feature enhancement method, FERD, achieves higher detection accuracy than AETNet while retaining low inference time. The method proposed in the present work, which incorporates both spectral and spatial information through complementary learning, secures further performance gains. Notably, this improvement comes with only a marginal increase in computational overhead. Specifically, the 100- and 200-band configurations employ Fusion (B), resulting in lower time costs. In contrast, the higher time cost for the 50-band setup is attributed to the use of Fusion (A), as it requires two separate detector passes. For more details on the fusion strategies, please refer to Table \ref{tab:fusion_on_HAD100}.

Figure~\ref{scores} visualizes the detection maps produced by various methods on the first 50 bands of the HAD100 test set, specifically highlighting four challenging scenes (Nos. 58, 76, 87, and 90). Visually, both AETNet and FERD enhance detection primarily by suppressing background noise compared to the baseline RX detector. The proposed method further amplifies this advantage, leading to even more effective noise suppression and clearer highlighting of anomalous regions.

\subsection{Comprehensive Analysis of Background Feature Enhancement}
In this subsection, we evaluate the effectiveness of our background feature enhancement strategy. The model used here is trained on the AVIRIS-Classic background dataset from the HAD100 benchmark. The trained model is then directly applied to the ABU scenes for background enhancement. Since the number of spectral bands in these scenes is relatively high, the feature fusion adopts strategy (B)—that is, the enhanced features from both spectral and spatial branches are summed and then passed to the anomaly detector.

\begin{table*}[htbp]
    \centering
    \caption{Performance Comparison of Feature Enhancement Methods across Different Scenes}
    \label{tab:fe}
    \footnotesize
    \setlength{\tabcolsep}{5pt}
    \renewcommand{\arraystretch}{1.25}
    \begin{tabularx}{\textwidth}{
        l
        *{3}{>{\centering\arraybackslash}X}
        *{3}{>{\centering\arraybackslash}X}
        *{3}{>{\centering\arraybackslash}X}
    }
        \toprule
        \multirow{2}{*}{\textbf{Scene}} &
        \multicolumn{3}{c}{\textbf{RX}} &
        \multicolumn{3}{c}{\textbf{AE}} &
        \multicolumn{3}{c}{\textbf{Auto-AD}} \\
        \cmidrule(lr){2-4} \cmidrule(lr){5-7} \cmidrule(lr){8-10}
        & Original & Enhanced & $\boldsymbol{\Delta}$ &
        Original & Enhanced & $\boldsymbol{\Delta}$ &
        Original & Enhanced & $\boldsymbol{\Delta}$ \\
        \midrule
        (a) Airport-1 & 0.8248 & \textbf{0.9515} & +0.1267 &
                       0.8530 & \textbf{0.8773} & +0.0243 &
                       0.6941 & \textbf{0.8723} & +0.1782 \\
        (b) Airport-2 & 0.8433 & \textbf{0.9871} & +0.1439 &
                       0.9260 & \textbf{0.9872} & +0.0612 &
                       0.6764 & \textbf{0.9407} & +0.2643 \\
        (c) Airport-3 & 0.9301 & \textbf{0.9715} & +0.0415 &
                       0.9429 & \textbf{0.9596} & +0.0167 &
                       0.9210 & \textbf{0.9327} & +0.0117 \\
        (d) Airport-4 & 0.9526 & \textbf{0.9935} & +0.0409 &
                       0.8478 & \textbf{0.9541} & +0.1062 &
                       0.5509 & \textbf{0.9486} & +0.3978 \\
        (e) Beach     & 0.9538 & \textbf{0.9888} & +0.0350 &
                       \textbf{0.8747} & 0.8283 & --0.0464 &
                       \textbf{0.9832} & 0.5000 & --0.4832 \\
        (f) Urban     & 0.9513 & \textbf{0.9756} & +0.0243 &
                       0.8662 & \textbf{0.9734} & +0.1072 &
                       0.7663 & \textbf{0.9735} & +0.2072 \\
        \midrule
        \textbf{Mean Time (s)} &
        \textbf{0.13} & 0.29 & --- &
        \textbf{1.51} & 1.65 & --- &
        \textbf{9.23} & 9.38 & --- \\
        \bottomrule
    \end{tabularx}
\end{table*}

Table~\ref{tab:fe} presents the performance evaluation of our background feature enhancement method applied to Mahalanobis distance-based (RX) and autoencoder-based (AE, Auto-AD) anomaly detection algorithms across six real hyperspectral scenes. The results demonstrate that the proposed enhancement significantly improves the detection accuracy of RX with minimal additional time cost. For AE and Auto-AD, consistent performance gains are observed in five out of six scenes, indicating the generality and robustness of our method.

\begin{figure*}[!htbp]
	\centering
	\includegraphics[scale=0.18]{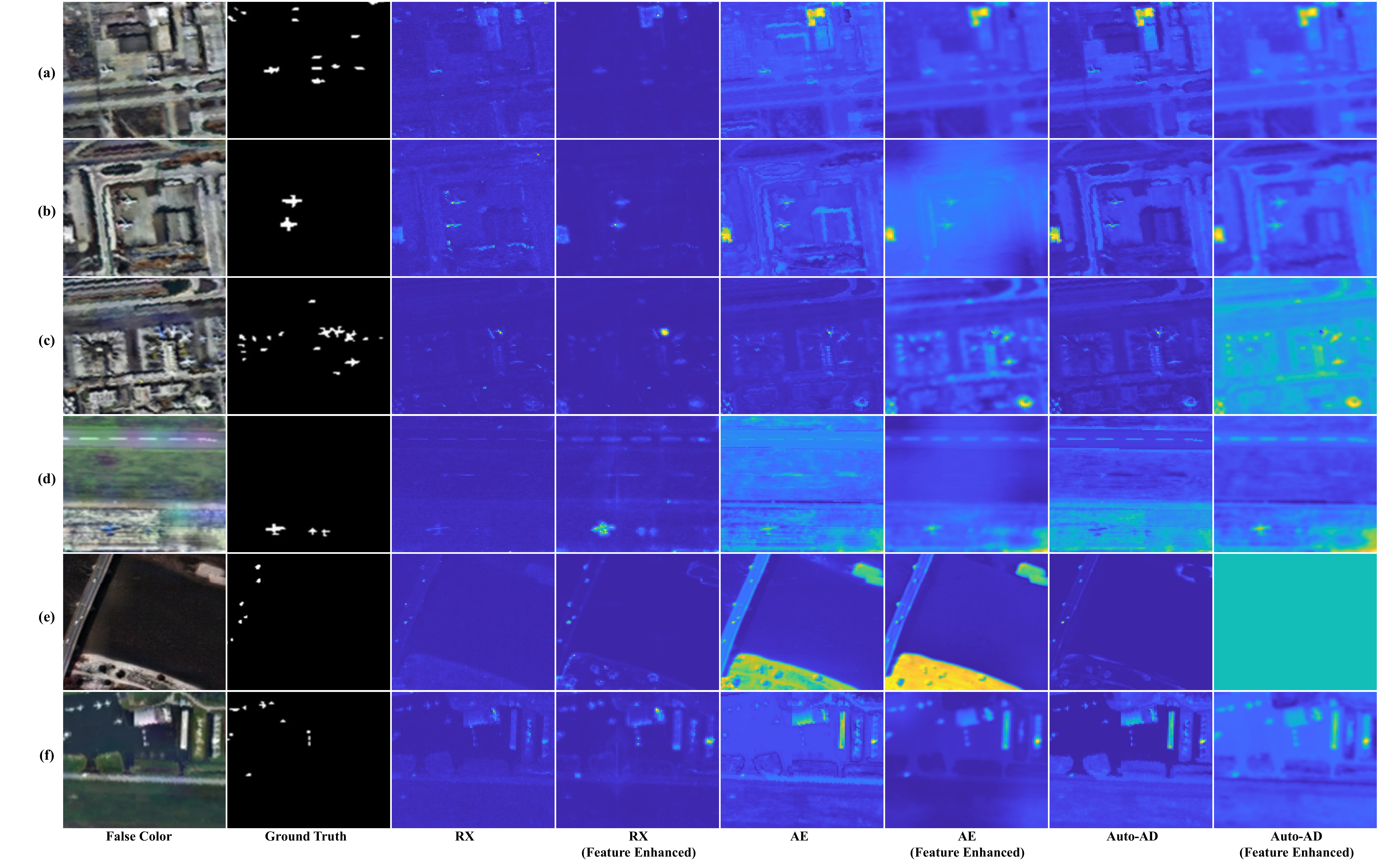}
	\caption{Detection maps comparing original and feature-enhanced results across six test scenes: (a) Airport-1, (b) Airport-2, (c) Airport-3, (d) Airport-4, (e) Beach, (f) Urban. Rows 3--4: RX detection maps (original vs. enhanced). Rows 5--6: AE detection maps (original vs. enhanced). Rows 7--8: Auto-AD detection maps (original vs. enhanced).}
	\label{result_fe}
\end{figure*}

The detection maps in Fig.~\ref{result_fe} provide visual evidence of the effectiveness of our enhancement strategy. In most cases, background interference is notably suppressed, while the anomaly regions exhibit higher saliency and clearer boundaries.

\begin{figure}[!htbp]
	\centering
	\includegraphics[scale=0.28]{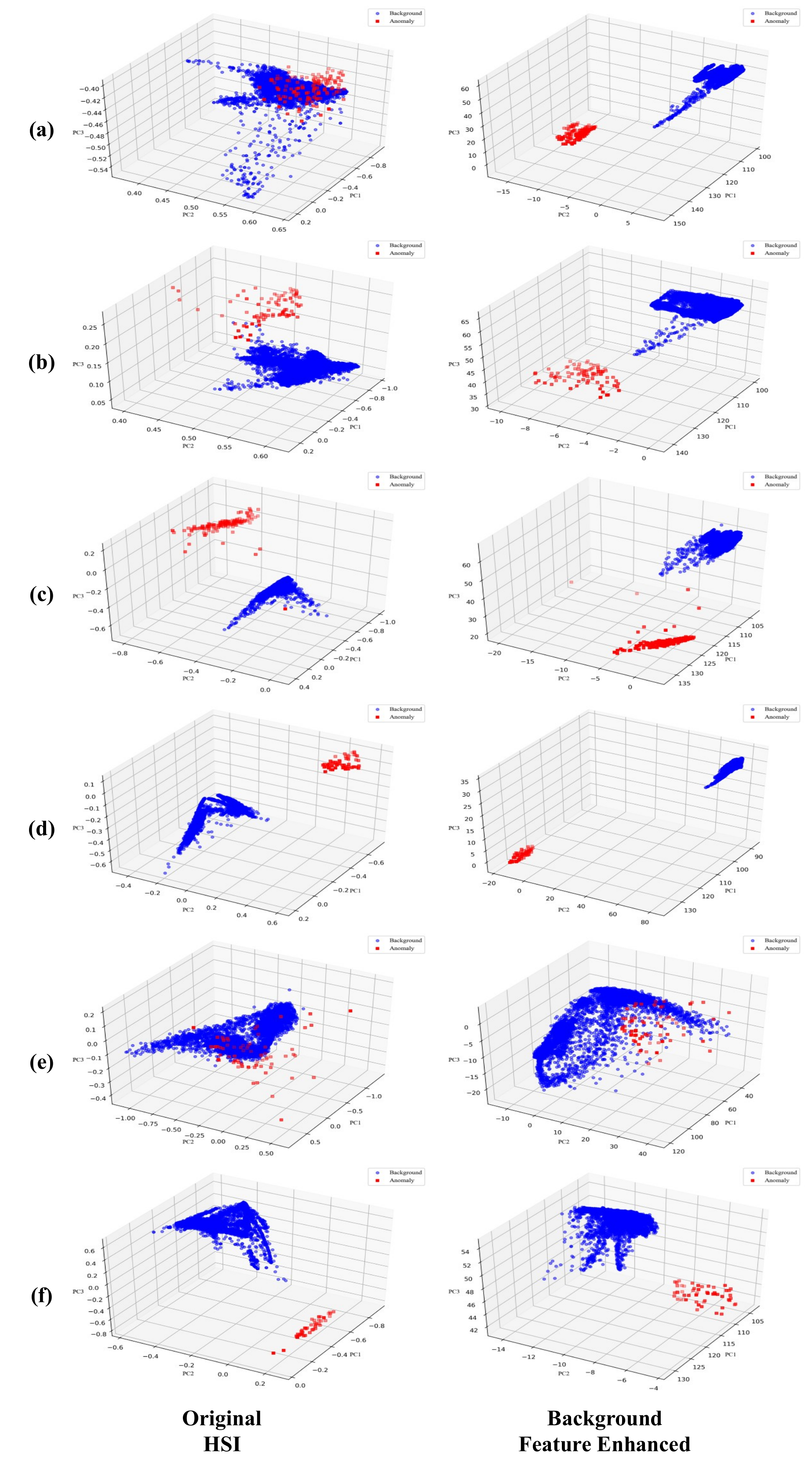}
	\caption{PCA visualization of spectral distributions: (left) original HSI data projected onto the first three principal components; (right) feature-enhanced HSI data projection. Blue points denote background pixels, while red points indicate anomalous pixels.}
	\label{pca}
\end{figure}

Fig.~\ref{pca} further illustrates the distributional changes before and after feature enhancement through PCA-reduced visualizations, where background and anomaly pixels are separately projected onto their respective principal component spaces for clearer observation of their distribution characteristics. Each point corresponds to a pixel projected onto the first three principal components. After enhancement, background pixels exhibit distributions that more closely approximate Gaussian forms, while anomaly pixels become more separable in most scenes. From these observations, several key insights can be summarized as follows:

\begin{enumerate}
    \item \textbf{RX sensitivity to Gaussianity.}  
    RX assumes a multivariate Gaussian background model. As shown from scenes (a) to (f), after enhancement, all HSIs exhibit distributions that more closely resemble Gaussian ellipsoids rather than irregular scatterings. Consequently, RX achieves consistent performance improvements across all scenes. Notably, scene (e) (“Beach”) demonstrates that even when anomaly–background separability remains low, a more Gaussian-like distribution still enhances RX performance.

    \item \textbf{RX sensitivity to separability.}  
    RX performance is also influenced by the separability between anomaly and background pixels. For example, in scenes (a) and (b), the original data show poor separability and highly non-Gaussian distributions, leading to suboptimal detection. After enhancement, both Gaussian conformity and separability improve substantially, resulting in the largest accuracy gains among all scenes.

	\item \textbf{AE and Auto-AD sensitivity to separability.}  
	For AE and Auto-AD detectors, the anomaly–background separability (distance) is the primary factor affecting performance. In scenes (a), (b) and (d), where separability increases notably after enhancement, detection performance improves substantially. By contrast, scene (c) shows only minor changes in separability and correspondingly little performance gain.

    \item \textbf{AE and Auto-AD sensitivity to distributional regularity.}  
	As observed in scene (f), even with limited improvement in separability, both AE and Auto-AD achieve better detection results after feature enhancement, primarily due to the more Gaussian-like background distributions. Theoretically, both methods reconstruct the input data to learn its underlying distribution. When the background features become more regular (e.g., Gaussianized), the reconstruction process better captures the intrinsic background structures, thereby reducing reconstruction noise and improving anomaly localization accuracy.

	\item \textbf{Scene (e) presents an interesting anomaly.}  
	Although the background distribution becomes slightly more Gaussian after enhancement, the overall separability between anomalies and background remains extremely low. Consequently, the detection performance of AE and Auto-AD decreases—particularly for Auto-AD, where anomalies become nearly indistinguishable. This phenomenon suggests that certain latent factors beyond Gaussianity and separability may also play a role in influencing AE and Auto-AD performance, and our feature enhancement network appears to suppress these factors. It is also possible that Gaussianization is only beneficial when a certain degree of anomaly–background separability already exists, as observed in scene (f).
\end{enumerate}

Overall, both quantitative and qualitative analyses confirm that the proposed feature enhancement effectively reshapes the spectral feature space toward Gaussian-like background distributions and more discriminative anomaly structures, thereby benefiting both statistical and learning-based anomaly detectors.

\subsection{Ablation Study}

\subsubsection{Effect of Individual Branches and Fusion Strategies}

\begin{table*}[htbp]
	\centering
	\caption{erformance of individual branches and two fusion strategies on the HAD100 test set. The lighter rows correspond to single-branch background enhancement, while the darker rows show the results of spectral–spatial fusion.}
	\renewcommand{\arraystretch}{1.1} 
	\begin{adjustbox}{max width=1\linewidth}
		\begin{tabular}{l *{6}{c}}
			\toprule
			\multirow{2}{*}{Method} & \multicolumn{2}{c}{The First 50 Bands} & \multicolumn{2}{c}{The First 100 Bands} & \multicolumn{2}{c}{The First 200 Bands} \\
			\cmidrule(lr){2-3} \cmidrule(lr){4-5} \cmidrule(lr){6-7}
			& \multicolumn{1}{c}{\textbf{mAUC}} & \multicolumn{1}{c}{\textbf{Time}} & \multicolumn{1}{c}{\textbf{mAUC}} & \multicolumn{1}{c}{\textbf{Time}} & \multicolumn{1}{c}{\textbf{mAUC}} & \multicolumn{1}{c}{\textbf{Time}} \\
			\midrule
			Without Enhanced (RX)   & 0.97990 & \bfseries 0.017 & 0.97140 & \bfseries 0.024 & 0.96490 & \bfseries 0.052 \\

			\rowcolor{lightgray}
			Spectral Enhanced (FERD)  &  0.99408 & 0.037 &  0.99010 & 0.044 &  0.98906 & 0.073 \\
			\rowcolor{lightgray}
			Spatial Enhanced  & 0.97943 & 0.036 & 0.97131 & 0.043 & 0.96547 & 0.071 \\
			\midrule

			\rowcolor{darkgray}
			Spectral-Spatial Fusion (A) & \bfseries 0.99530 &  0.071 &  0.99008 &  0.079 &  0.98061 & 0.142 \\
			\rowcolor{darkgray}
			Spectral-Spatial Fusion (B) &  0.99386 &  0.041 &  \bfseries 0.99102 &  0.046 &  \bfseries 0.99021	& 0.075 \\
			\bottomrule
		\end{tabular}%
	\end{adjustbox}
	\label{tab:fusion_on_HAD100}%
\end{table*}%

Table~\ref{tab:fusion_on_HAD100} presents the performance comparison of different branches and fusion strategies on the HAD100 test set. Results with lighter shading correspond to single-branch background enhancement, where the spectral enhancement branch (FERD) represents our previous method, and the spatial enhancement branch corresponds to enhancement using only the spatial branch. The darker rows show the results of spectral–spatial fusion using two different fusion strategies (A) and (B), as described in Section~\ref{method_fusion}.

As shown in the table, applying background feature enhancement on the spectral branch alone (i.e., FERD) already brings substantial improvement with minimal additional computational cost. In contrast, the spatial enhancement branch performs poorly when used independently, even slightly worse than the non-enhanced baseline. This is expected, as the spatial branch is designed to complement rather than replace the spectral branch, focusing on spatial cues that the spectral branch may overlook. This further confirms that it primarily serves as a complementary learner rather than a standalone detector, whereas the spectral branch is independently trainable and capable of effective detection on its own.

When the spectral and spatial branches are fused, the overall performance shows a substantial improvement over any single-branch configuration, validating the effectiveness of the proposed complementary learning paradigm. This demonstrates that the spatial branch indeed captures informative cues that the spectral branch alone cannot learn. The two fusion strategies behave differently depending on the scenario: Fusion (A) is a decision-level fusion that multiplies the anomaly maps from both branches, which emphasizes spatial consistency and tends to be more effective when the number of spectral bands is relatively small. Fusion (B), on the other hand, performs feature-level fusion within the spectral feature space and becomes more beneficial as the number of bands increases, leveraging richer spectral representations.

\subsubsection{Effect of Different Loss Components}

\begin{table}[htbp]
	\centering
	\caption{Ablation study on loss functions in Stage 1 (Spectral Branch)}
	\renewcommand{\arraystretch}{1.1} 
	\begin{tabularx}{\linewidth}{*{4}{>{\centering\arraybackslash}X}} 
		\toprule
		$\mathcal{L}_{\text{Z}}$  & $\mathcal{L}_{\text{mse}}$  & $\mathcal{L}_{\text{sim}}$  & \textbf{Mean AUC} \\
		\midrule
		$\checkmark$   &  &  & 0.9913 \\
		& $\checkmark$   & $\checkmark$   & 0.9934 \\
		$\checkmark$   & $\checkmark$   &  & 0.9917 \\
		$\checkmark$   &  & $\checkmark$   & 0.9924 \\
		\cmidrule(lr){1-4} 
		$\checkmark$   & $\checkmark$   & $\checkmark$   & \textbf{0.9941} \\
		\bottomrule
	\end{tabularx}%
	\label{tab:Loss_step1}%
\end{table}%

Table \ref{tab:Loss_step1} presents the ablation study of the loss components used in Stage 1 training of the spectral branch, evaluated on the first 50 bands of the HAD100 dataset.
The results show that both the spectral alignment loss ($\mathcal{L}_{\text{Z}}$) and the reconstruction losses—$\mathcal{L}_{\text{mse}}$ for input reconstruction and $\mathcal{L}_{\text{sim}}$ for multi-scale feature reconstruction—each make positive contributions to the model’s performance, but none alone achieves optimal results.
Similarly, any combination missing one of these three loss terms leads to a noticeable decline in detection accuracy.
This demonstrates that both the spectral alignment mechanism and the full reconstruction process are indispensable, as omitting any component compromises the overall detection capability.
The highest mean AUC (0.9941) is obtained only when all three losses are jointly optimized, confirming that the alignment and reconstruction mechanisms operate synergistically to enable the teacher network to learn robust and discriminative spectral representations.

\begin{table}[htbp]
	\centering
	\caption{Ablation study on loss functions in Stage 2 (Spatial Branch)}
	\renewcommand{\arraystretch}{1.1}
	\setlength{\tabcolsep}{3pt} 
	\begin{tabularx}{\linewidth}{*{5}{>{\centering\arraybackslash}X}}
		\toprule
		$\mathcal{L}_{\text{cc}}$  & $\mathcal{L}_{\text{cos}}$  & $\mathcal{L}_{\text{var}}$  & $\mathcal{L}_{\text{recon}}$ & \textbf{Mean AUC} \\
		\midrule
		$\checkmark$ & $\checkmark$ & $\checkmark$ & & 0.9945 \\
		& & & $\checkmark$ & 0.9947 \\
		$\checkmark$ & $\checkmark$ & & $\checkmark$ & 0.9947 \\
		$\checkmark$ & & $\checkmark$ & $\checkmark$ & 0.9951 \\
		& $\checkmark$ & $\checkmark$ & $\checkmark$ & 0.9950 \\
		\cmidrule(lr){1-5}
		$\checkmark$ & $\checkmark$ & $\checkmark$ & $\checkmark$ & \textbf{0.9953} \\
		\bottomrule
	\end{tabularx}
	\label{tab:Loss_step2}
\end{table}

Table \ref{tab:Loss_step2} presents an ablation study on the loss components employed during Stage 2 training of the spatial branch, evaluated on the first 50 bands of the HAD100 dataset. Several key observations emerge from the results.
First, when only the decorrelation losses ($\mathcal{L}_{\text{cc}}$, $\mathcal{L}_{\text{cos}}$, $\mathcal{L}_{\text{var}}$) are used without the reconstruction loss $\mathcal{L}_{\text{recon}}$, the performance is poor, confirming that without reconstruction the student branch tends to learn meaningless noise instead of useful features.
Second, using $\mathcal{L}_{\text{recon}}$ alone gives a decent result, showing that reconstruction is crucial for learning meaningful representations, but it cannot guarantee complementarity to the spectral teacher.
Third, removing $\mathcal{L}_{\text{var}}$ leads to a noticeable drop in performance, indicating that variance preservation is essential for preventing information collapse.
Finally, the best performance is achieved when all losses are combined. This demonstrates that $\mathcal{L}_{\text{cc}}$ and $\mathcal{L}_{\text{cos}}$ help learn different types of complementary information, while $\mathcal{L}_{\text{var}}$ and $\mathcal{L}_{\text{recon}}$ jointly ensure stable and meaningful feature learning.

\section{CONCLUSION}
In this paper, we present a novel \textbf{``Rebellious Student''} framework for complementary spectral–spatial background feature enhancement in hyperspectral anomaly detection. Our method leverages a two-stage strategy: a spectral enhancement network captures robust background spectral representations via reverse distillation, while a spatial branch—the “rebellious student”—is trained with a feature decorrelation loss to learn complementary spatial patterns absent from the spectral branch. Experimental results on the HAD100 benchmark demonstrate that incorporating spatial information yields substantial improvements over spectral-only enhancement (FERD), with only marginal computational overhead. Cross-scene evaluations on the ABU dataset further reveal that the learned enhancement mechanism reshapes features toward more Gaussian-like and discriminative structures, which underpins the improved performance of both statistical and learning-based anomaly detectors.

However, our study also reveals an interesting limitation worth further investigation. As observed in Scene (e), while our feature enhancement generally promotes more Gaussian-like background distributions, the performance of autoencoder-based detectors (AE and Auto-AD) may degrade when the inherent anomaly-background separability remains extremely low. This phenomenon suggests that beyond distribution Gaussianity and feature separability, other latent factors may influence these detectors' performance—factors that our current enhancement approach might inadvertently suppress. Looking forward, several promising research directions emerge from this work. First, the development of more adaptive enhancement strategies is a promising direction, particularly for handling challenging scenarios with low inherent separability where current approaches show limitations. Second, the complementary learning paradigm could potentially be extended to other remote sensing tasks beyond anomaly detection—such as classification, segmentation, or change detection. More broadly, the core concept of the ``Rebellious Student'' framework — which enables the model to strategically learn divergent yet complementary representations — may be applicable to other domains that require feature disentanglement or multi-modal learning.

\end{document}